\documentclass[12pt]{article}

\usepackage{graphicx}
\usepackage{authblk}
\usepackage{lmodern}
\usepackage[T1]{fontenc}
\usepackage[utf8]{inputenc}
\usepackage[margin=1in]{geometry}
\usepackage{amssymb,amsmath,amsthm,amsfonts}
\usepackage{float}
\usepackage{tabularx}

\usepackage{xcolor}

\usepackage{fancyhdr}
\usepackage{booktabs}
\begin{document}
%

%
%
%
\title{Explainable Face Recognition via Improved Localization}
%
%
%
\newcommand{\clrksn}{{*}}
\newcommand{\rit}{{$^\mathsection$}}
\newcommand{\afrl}{{$^{\ddagger}$}}

\renewcommand*{\Authsep}{\ }
\renewcommand*{\Authand}{\ }
\renewcommand*{\Authands}{\ }

\author[\clrksn]{Rashik Shadman}
\author[\rit]{Daqing Hou}
\author[\clrksn]{Faraz Hussain}
\author[\afrl]{M G Sarwar Murshed} 

\affil[\clrksn]{Clarkson University, Potsdam, NY, USA\authorcr {\tt \{shadmar,fhussain\}@clarkson.edu}\vspace{0.4em}}
\affil[\rit]{Rochester Institute of Technology, Rochester, NY, USA\authorcr{\tt dqvse@rit.edu}\vspace{0.4em}}
\affil[\afrl]{University of Wisconsin-Green Bay, Green Bay, WI, USA\authorcr{\tt murshedm@uwgb.edu}\vspace{0.4em}}

\maketitle

\begin{abstract}
Biometric authentication has become one of the most widely used tools in the current technological era to authenticate users and to distinguish between genuine users and imposters. Face is the most common form of biometric modality that has proven effective.
Deep learning-based face recognition systems are now commonly used across different domains. 
However, these systems usually operate like black-box models that do not provide necessary explanations or justifications for their decisions. This is a major disadvantage because users cannot trust such artificial intelligence-based biometric systems and may not feel comfortable using them when clear explanations or justifications are not provided. This paper addresses this problem by applying an efficient method for explainable face recognition systems. We use a Class Activation Mapping (CAM)-based discriminative localization (very narrow/specific localization) technique called Scaled Directed Divergence (SDD) to visually explain the results of deep learning-based face recognition systems. We perform fine localization of the face features relevant to the deep learning model for its prediction/decision. Our experiments show that the SDD Class Activation Map (CAM) highlights the relevant face features very specifically compared to the traditional CAM and very accurately.
The provided visual explanations with narrow localization of relevant features can ensure much-needed transparency and trust for deep learning-based face recognition systems.
\end{abstract}

\section*{Keywords}
Explainable AI; cybersecurity; biometrics; localization; class activation mapping.

%
\maketitle

\section{Introduction} \label{intro}
Deep learning models have become ubiquitous across a wide range of application domains due to their efficiency and high performance. The field of biometric authentication is also adopting deep learning-based systems such as face, fingerprint, and iris recognition systems. However, in most cases, the results or decisions of these models are not transparent, i.e., it is not clear why a deep learning model made a specific prediction. This can lead to a lack of trust in the results. To achieve transparency and increase the trustworthiness of such systems, \emph{explainable AI} techniques have drawn much attention \cite{R0}. Explainability seeks to provide the reason behind the prediction/classification/decision of a deep learning model. There are multiple types of explanation techniques; however, this paper primarily focuses on the visual explanations of deep learning models \cite{R01} which is called \emph{discriminative localization}\footnote{Discriminative localization refers to the process of identifying and highlighting specific regions in an image that are most relevant or important for a model's decision-making, particularly for distinguishing between different classes} \cite{R4}.

Explainability can enhance trust in deep learning-based biometric systems such as face recognition. Explaining the predictions of deep learning models is a complicated task, as the structure of a deep learning model is very complex \cite{R1}. There are different techniques to visually explain the decision of a deep learning model, such as Saliency Maps \cite{R2}, LIME \cite{R3}, Class Activation Mapping (CAM) \cite{R4}, Gradient-weighted Class Activation Mapping (Grad-CAM) \cite{R5} etc.

In this paper, we use a CAM-based discriminative localization technique Scaled Directed Divergence (SDD) \cite{R6} for narrow/specific localization of relevant (relevant to the deep learning model for its prediction/decision) spatial regions of face images. A Class Activation Map (CAM) highlights the class-specific discriminative regions relevant to the deep learning model for arriving at its prediction/decision. The SDD technique is particularly useful for the fine localization of class-specific features in situations of class overlapping. Class overlapping occurs when images from different classes share similar regions or features, making it challenging for a model to distinguish between them. Therefore, the SDD technique is suitable for visually explaining the decisions of face recognition systems, which often deal with similar face images (face images of different individuals that share similar features).

In a previous paper, Williford et al. \cite{R7} performed explainable face recognition using a new evaluation protocol called the ``Inpainting Game''. Given a set of three images (probe, mate 1, and nonmate), the explainable face recognition algorithm was to identify pixels in a specific region that contribute to the mate's identity more than to the nonmate's identity. This identification was represented through a saliency map, indicating the discriminative regions for mate recognition. The ``Inpainting Game" protocol was used for the evaluation of the identification made by the explainable face recognition algorithm. However, Williford et al. did not consider a multiclass model for explainable face recognition where there are both similar and nonsimilar classes. Rajpal et al. \cite{R8} used LIME to visually explain the predictions of face recognition models. However, they did not consider the case of class overlap (common areas highlighted in the visual explanations of different classes), therefore, their explanations failed to narrowly show the important features responsible for making a decision.

In our work, we train a convolutional neural network model with a multiclass face image dataset (FaceScrub dataset \cite{R14}) and compute the test accuracy. Then, for a given test image, we check the predicted class of the model and perform fine localization of the most relevant spatial regions (relevant to the model for its prediction) by implementing the SDD technique. The output of the SDD technique is represented as the SDD CAM, where finer localization of the most relevant features is shown compared to the traditional CAM. The methodology, algorithm, experimental setup, results, and discussion are described in sections \ref{CAM}, \ref{XAI}, \ref{evaluation}, and \ref{discuss}.

\section{Related Work}\label{rw}
In this section, we describe some of the most prominent visual explanation techniques. The first technique is the Saliency Maps proposed by Simonyan et al. \cite{R2}. It is an effort to prioritize pixels within an input image based on their significance in influencing the output score within a Convolutional Neural Network (CNN). In CNNs, the output score exhibits nonlinear behavior. To approximate the model's response in the vicinity of an input image, a first-order Taylor expansion is employed. This expansion results in an equation akin to a linear model. The determination of weight values, crucial for this approximation, is achieved through the process of backpropagation. Subsequently, the Saliency Map is derived by reconfiguring the weight vector to conform to the spatial dimensions of the input image.

Ribeiro et al. \cite{R3} proposed a novel explanation technique, Local Interpretable Model-agnostic Explanations (LIME), which operates by learning an interpretable model locally around the prediction, thus explaining the predictions of any classifier. This interpretable model is locally trained for specific input instances and approximates the prediction. So, LIME helps break down why a complex model made a certain prediction for a single input instance by creating a simpler explanation around that prediction.

Deep Learning Important Features (DeepLIFT) \cite{R9} is another explainability method. It is employed to decompose the output prediction generated by a neural network concerning a particular input. This decomposition is facilitated by the backpropagation of the contributions of all neurons within the network to each feature of the input. DeepLIFT introduces a comparative analysis by evaluating the activation of individual neurons in relation to their respective reference activation and assigns contribution scores based on the difference. The computation of these scores is efficiently executed within a single backward pass.

Springenberg et al. \cite{R10} proposed the Guided Backpropagation technique, which is a combination of "deconvnet" and "backpropagation" methods. In this approach, an extra guiding signal is incorporated from higher layers into the standard backpropagation process. This effectively blocks the reverse propagation of negative gradients, which represent the neurons responsible for reducing the activation of the targeted higher-layer unit that we intend to visualize. Guided Backpropagation enables the visualization of not only the final layers but also intermediate layers within the network.

Class Activation Mapping \cite{R4} is a very effective visual explanation technique. A Class Activation Map (CAM) of a specific class serves to delineate the discerning regions within an image that a CNN utilizes in the identification of said class. The architecture predominantly comprises convolutional layers, and in the proximity of the final output layer, specifically the softmax layer in categorical classification scenarios, a global average pooling operation is conducted on the convolutional feature maps. These pooled features subsequently serve as input to a fully connected layer, responsible for generating the ultimate output, be it categorical or otherwise.
The discernment of the relative importance of image regions is achieved by back-projecting the weights of the output layer onto the convolutional feature maps. The resulting CAM effectively highlights the regions within the image that are discriminative and class-specific in nature. The details of CAM generation are described later in Section \ref{CAM}. The Scaled Directed Divergence (SDD) technique is based on CAMs \cite{R6} and is extensively used in this work to visually explain face recognition (Section \ref{XAI}).

Gradient-weighted Class Activation Mapping (Grad-CAM) \cite{R5} is a generalization of CAM. Grad-CAM leverages the gradients associated with a designated target concept, such as the logits corresponding to the classification of `dog' or even a descriptive caption. These gradients are traced back through the network to the final convolutional layer, facilitating the generation of a coarse localization map.
This map effectively highlights crucial regions within the image that play a significant role in predicting the specified concept. The neurons within the ultimate convolutional layers specifically seek semantic, class-specific information within the image, such as distinct object parts. By utilizing the gradient information streaming into the last convolutional layer of the CNN, Grad-CAM discerns the significance of each neuron in relation to a particular decision of interest. This method is notably characterized by its high degree of class discrimination. Also, Selvaraju et al. \cite{R5} proposed Guided Grad-CAM, which is a combination of Grad-CAM and Guided Backpropagation. Guided Grad-CAM provides high-resolution class-discriminative visualizations.

\begin{figure}[hbt!]
\centering
\includegraphics[width=0.8\textwidth, height=11cm]{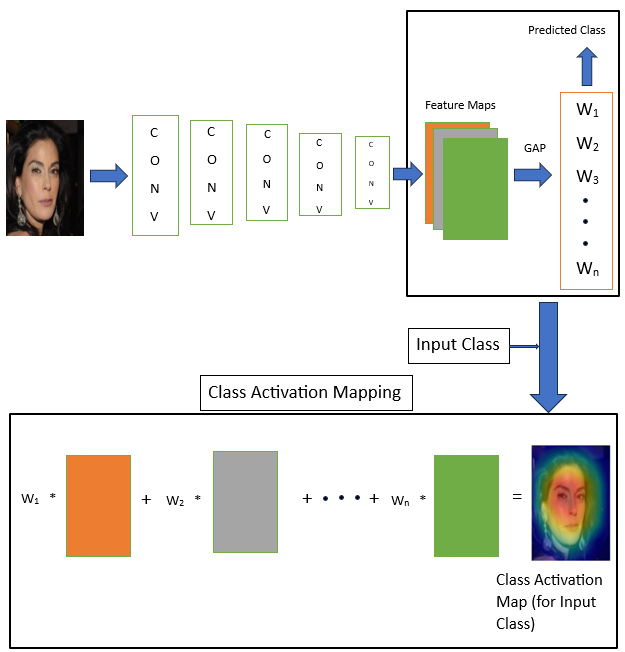}
\caption{Steps of generating CAMs for visual explainability \cite{R4}. The most important layers in CNN are convolutional layers, which capture significant features of the input image. The feature maps of the last convolutional layer are global average pooled before the output layer. The weights of the output layer are mapped back to the convolutional feature maps (last convolutional layer) to generate the CAM for the input class. The CAM is a weighted sum of the convolutional feature maps. \label{cam}}
\end{figure}   
\unskip

\section{Methodology and Algorithm} \label{CAM}
In this section, we describe the Class Activation Maps (CAMs) \cite{R4} and Scaled Directed Divergence (SDD) \cite{R6}. We use the SDD technique for our experiments on visual explainability. The SDD technique uses CAMs and is suitable for scenarios with overlapping classes, which is mentioned in section \ref{intro}.

\subsection{Class Activation Maps}
The main goal of Class Activation Maps (CAMs) is to highlight the regions of an image that are most influential for the decision of a CNN model. CAMs are useful for visually explaining the model's focus areas.

This approach involves tracing the predicted class score back to the preceding convolutional layer to generate CAMs. These maps serve to accentuate regions in the input image that are specifically relevant to the identified class, showcasing discriminative features.

The last convolutional layer in CNN captures features significant for the prediction of the CNN model. These features are used to generate CAMs for the respective classes. The weighted sum of the feature maps is computed to generate CAMs. The CAM generation procedure is explained in Fig. \ref{cam}.

Following the terminology and notations from Zhou et al. \cite{R4}, let softmax be the final output layer. For a test image, $f_k(x,y)$ is the activation of unit $k$ at spatial location $(x,y)$ in the last convolutional layer. Then, global average pooled result $F_k$ for unit $k$ is $\sum_{x,y}f_k(x,y)$. The input to the softmax ($S_c$) for target class \footnote{The target class refers to the specific class label for which the model's attention or activation map is generated, highlighting the image regions that contribute most to the model's prediction of that class.} $c$ is $\sum_{k}w_{k}^{c}F_{k}$. Here, $w_{k}^{c}$ represents the weight corresponding to class $c$ for unit $k$. Basically, the significance of $F_k$ for class $c$ is recognized by $w_{k}^{c}$. From $F_k = \sum_{x,y}f_k(x,y)$, the class score $S_c$ can be written as:
\begin{equation}
    S_c = \sum_{k}w_{k}^{c}\sum_{x,y}f_k(x,y) = \sum_{x,y}\sum_{k}w_{k}^{c}f_k(x,y)
\end{equation}
The CAM $M_c$ for class $c$ is-
\begin{equation}
   M_c(x,y) = \sum_{k}w_{k}^{c}f_k(x,y)
\end{equation}
Here, $M_c(x,y)$ is the spatial element at grid $(x,y)$.

\begin{figure}[hbt!]
\includegraphics[width=\textwidth]{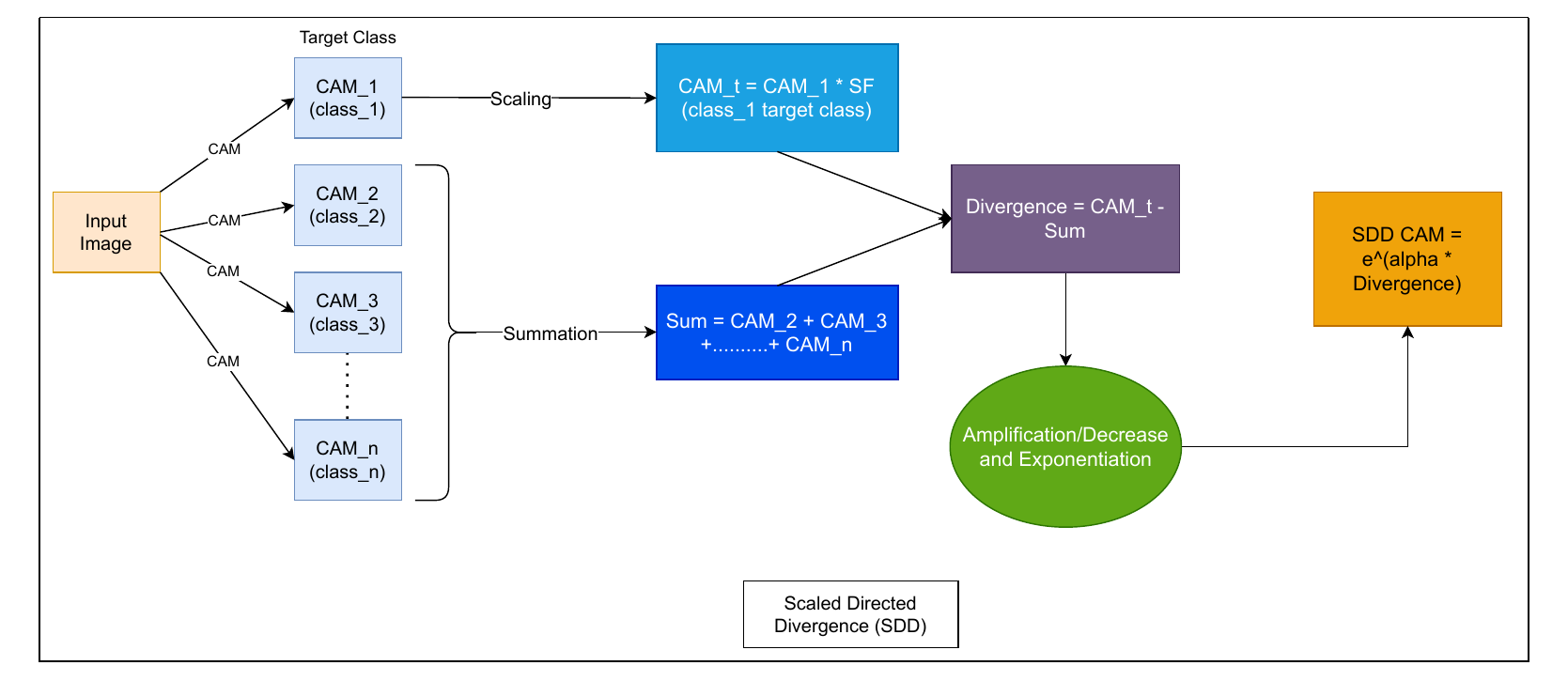}
\caption{An overview of the Scaled Directed Divergence (SDD) technique \cite{R6}. For an input image, multiple CAMs are generated for multiple classes that are overlaid on the same test image. Among the classes for which CAMs are generated, one class is selected as the target class (the selection of the target class is up to the user; it can be the original class, the predicted class, or any other class). \textbf{Here, we consider class\_1 as the target class.} As I mentioned, the target class is not fixed, and any class can be considered the target class. The target class CAM is multiplied by the scaling factor (SF) to adjust for differences in magnitude between CAMs to enable meaningful subtraction. Other CAMs are subtracted from the scaled target class CAM to generate the Divergence values. The Divergence values are multiplied by alpha (explained in section \ref{SDD}) for amplification/decrease before exponentiation. Finally, the SDD CAM is generated for the target class.\label{sdd}}
\end{figure}   
\unskip

\subsection{Scaled Directed Divergence (SDD)} \label{SDD}
Scaled Directed Divergence (SDD) \cite{R6} is a visual explainability technique based on CAMs. This technique is useful for the narrow/specific localization of relevant features (important to the CNN model for its prediction) of an input image compared to traditional CAMs. 

Considering a multi-class model of $i+1$ classes, let $S = \{x_o,.....,x_i\}$ be the set of CAMs. Let $T(t) = S\verb|\|\{x_t\}$ be the set of CAMs where the target class (the class for which SDD CAM is generated) CAM $x_t$ is removed from set $S$. Then, the SDD CAM for target class t is defined as:
\begin{equation}
    CAM_{SDD}^{t} = e^{\alpha(x_tF-\sum_{k\epsilon T(t)}x_k)}
\end{equation}
We have improved the SDD technique by updating the scaling factor $F$ and parameter $\alpha$. The target class CAM is scaled by $F$ (where $F$ = $|S|$ * (average mean absolute value of all CAMs except target class CAM / mean absolute value of target class CAM); the minimum value of $F$ is $S$ and the maximum value is 50) before computing the divergence values between the target CAM and other CAMs. Parameter $\alpha$ (where $\alpha = 0.2 + (stddev(CAM_{SDD}) / (stddev(CAM_{SDD}) + 1)) * 0.2$) amplifies or decreases the divergence values to be exponentiated.  The set of CAMs $S$ only needs to contain the target CAM and CAMs with significant overlap. An overview of the SDD technique is shown in Fig. \ref{sdd}.

\section{Explainable Face Recognition}\label{XAI}
Our experiment is divided into two parts: 1) Using SDD for explaining the face recognition model's results, and 2) Evaluation of SDD results. We first describe the face recognition model and use SDD to visually explain its results (Section \ref{model-1}).

\subsection{Face Recognition Model} \label{model-1}
We use the state-of-the-art AdaFace model \cite{R12} for face recognition.
The AdaFace model is ResNet100, which is pre-trained on the WebFace12M \cite{R13} face dataset. For the face classification task, the pre-trained AdaFace model is fine-tuned on the publicly available FaceScrub dataset \cite{R14}, which consists of 43148 images of 530 classes (each person is a class). Among the 530 classes, there are 265 male classes (23216 images) and 265 female classes (19932 images). We divide the dataset into train, validation, and test sets with a ratio of 80/10/10. So there are 34099 training images, 4555 validation images, and 4494 test images. The AdamW optimizer is used for fine-tuning the model. The learning rate is 0.005, and the number of epochs is 20. StepLR learning rate scheduler is used where step\_size = 10 and gamma = 0.1. The overall test accuracy of the model is 96.33\%.

\begin{figure}[hbt!]
\includegraphics[width=\textwidth]{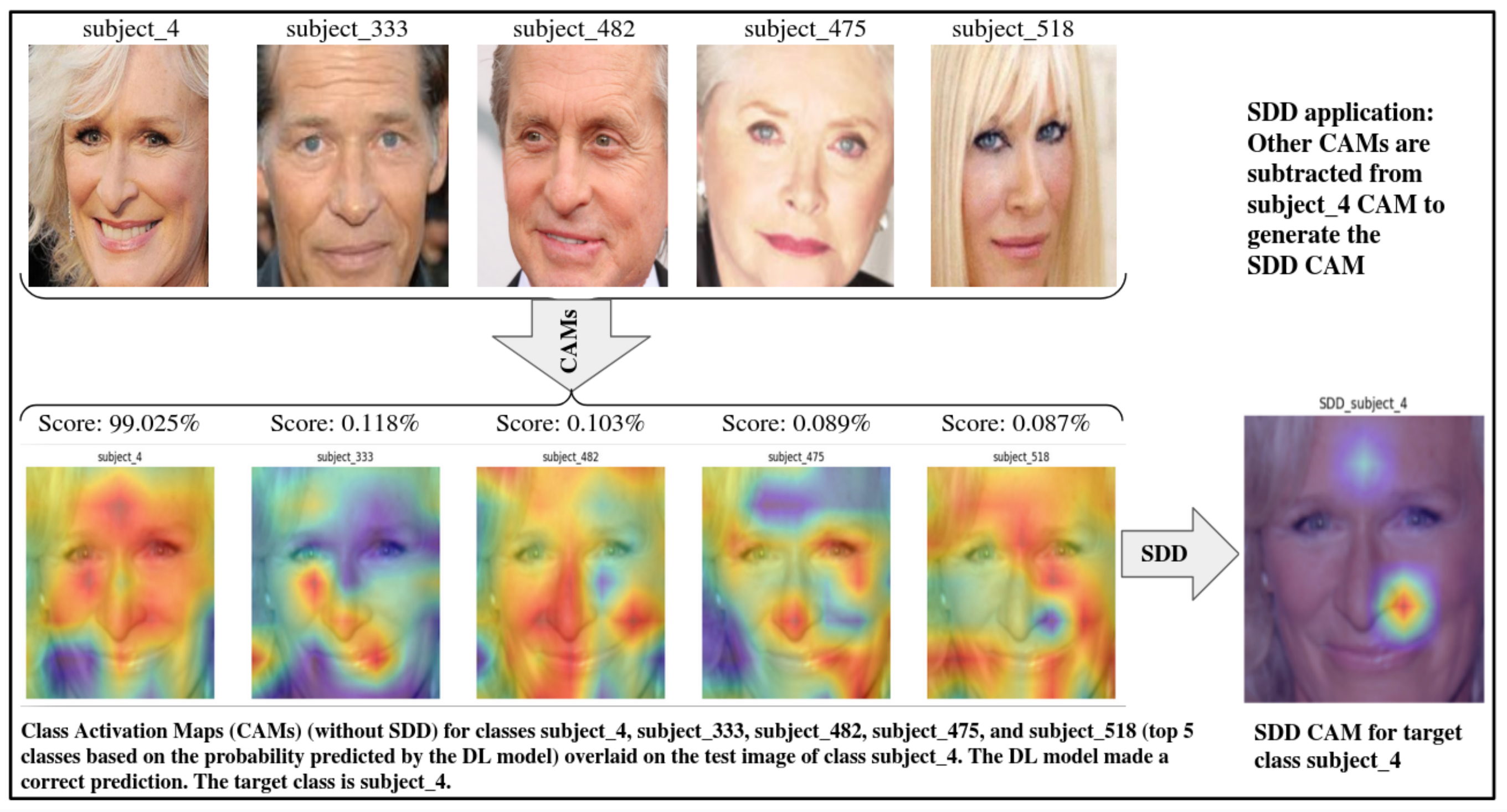}
\caption{Visually explaining the prediction of the face recognition model (described in section \ref{model-1}) by highlighting the relevant face features. The class of the test image is subject\_4. The DL model made a correct prediction of the test image. The probability of the top 5 classes predicted by the model: subject\_4 (99.025\%), subject\_333 (.118\%), subject\_482 (.103\%), subject\_475 (.089\%), subject\_518 (.087\%). In a CAM, the red, yellow, and green regions highlight the relevant features while the blue regions highlight the non-relevant features. The top row shows the example images of the classes.  The target class is subject\_4. In the bottom row, the first image (from the left) shows the traditional CAM for class subject\_4. The last image (SDD\_subject\_4) shows the SDD CAM for target class subject\_4, which localizes the most relevant features (\emph{left side area beside the nose and middle of the forehead}) in a very narrow manner (compared to the traditional CAM). Other CAMs are subtracted from the scaled target class CAM to generate the Divergence values, which are amplified/decreased and exponentiated to generate the SDD CAM. \label{fig1}}
\end{figure}   
\unskip

\subsection{Visual Explanation of Face Recognition}
Here, we visually explain the prediction of the face recognition model. We describe the fine localization of the most relevant face features (considered significant by the model), which visually explains the prediction of the CNN model. We consider 2 test images of 2 different classes (subject\_4 and subject\_120) that belong to the test set.

In Fig. \ref{fig1}, the class of the test image is subject\_4 which is predicted correctly by the face model. All the CAMs are overlaid on the test image of class subject\_4. The target class is subject\_4. For the SDD method, the top 5 classes (based on the probability predicted by the model) are considered. In the bottom row, the first 5 CAMs (from the left) are generated for the top 5 classes. The top row shows the sample images of the classes for which CAMs are generated. There is significant overlap among the CAMs of these classes. The last CAM (in the bottom row) is generated using SDD for target class subject\_4, in which very narrow/specific localization (compared to the traditional CAM of class subject\_4) of the most relevant face features is shown. The SDD CAM highlights the most important features of class subject\_4 (with respect to the other 4 classes) for the correct prediction of the model.

\begin{figure}[hbt!]
\includegraphics[width=\textwidth]{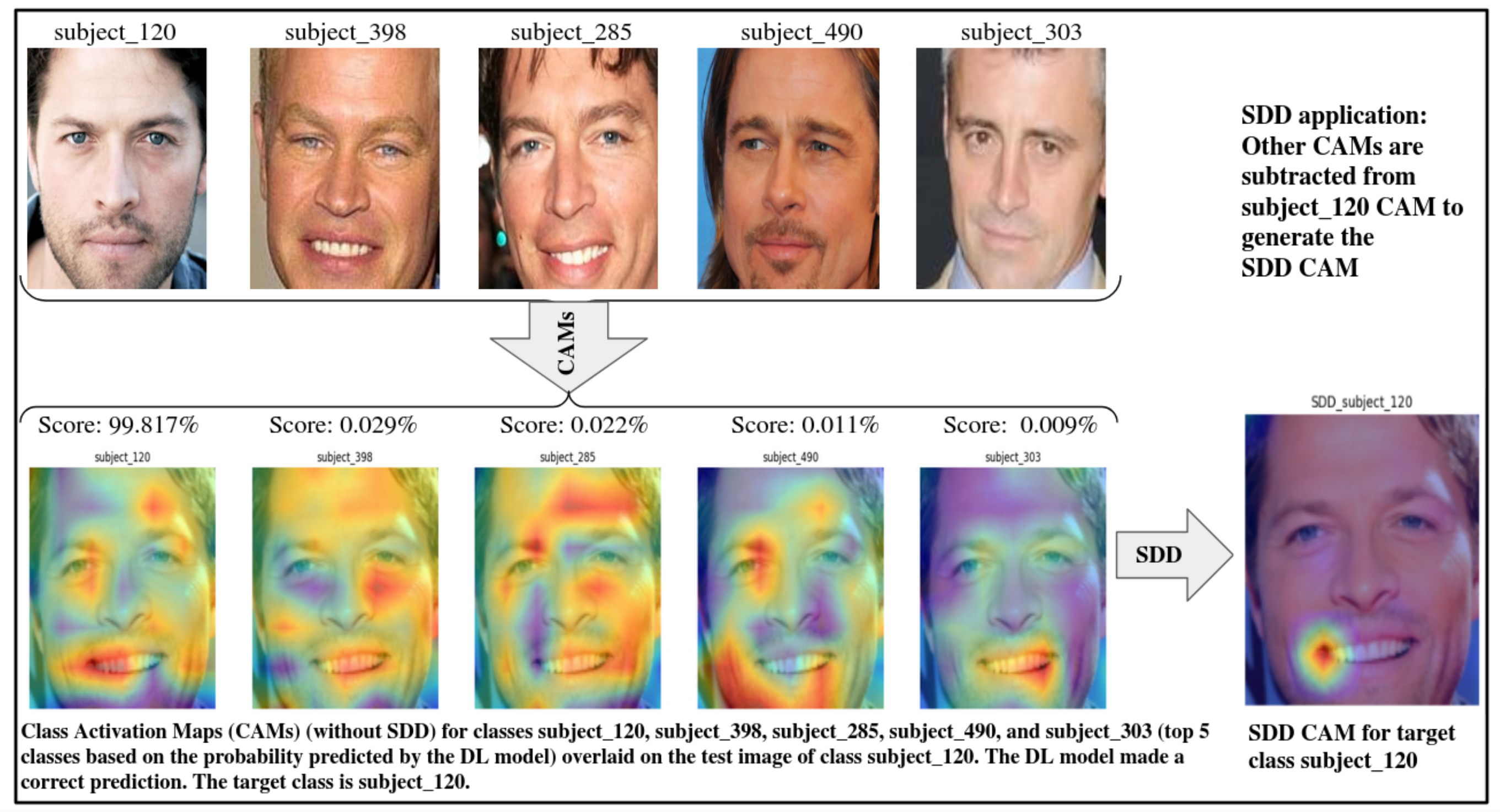}
\caption{Visually explaining the prediction of the face recognition model (described in section \ref{model-1}) by highlighting the relevant face features. The class of the test image is subject\_120. The DL model made a correct prediction of the test image. The target class is subject\_120. The probability of the top 5 classes predicted by the model: subject\_120 (99.817\%), subject\_398 (.029\%), subject\_285 (.022\%), subject\_490 (.011\%), subject\_303 (.009\%). The top row shows the example images of the classes. In the bottom row, the first image (from the left) shows the traditional CAM for class subject\_120. The last image  (SDD\_subject\_120) shows the SDD CAM for target class subject\_120, which localizes the most relevant features (\emph{right side area beside the lips}) in a very narrow manner (compared to the traditional CAM). Other CAMs are subtracted from the scaled target class CAM to generate the Divergence values, which are amplified/decreased and exponentiated to generate the SDD CAM. \label{fig2}}
\end{figure}   
\unskip

In Fig. \ref{fig2}, the class of the test image is subject\_120. The face model correctly predicted the class of the test image. Like the previous example, the CAMs are overlaid on the test image. The target class is subject\_120. Same as before, the top 5 classes are considered for implementing the SDD method. The last CAM (in the bottom row) is generated using SDD for target class subject\_120, in which very narrow/specific localization (compared to the traditional CAM of class subject\_120) of the most relevant face features is shown. The SDD CAM highlights the most important features of class subject\_120.

\section{Evaluation of the Visual Explanation}\label{evaluation}
In this section, we evaluate the visual explanations generated by the Scaled Directed Divergence method for the face recognition model. We use the deletion-and-retention evaluation scheme \cite{R15}. The deletion method evaluates the effectiveness of the visual explanations by removing the most important regions of an image identified by the explanation map and measuring the model's confidence drop (a higher drop is better). In the case of the retention method, evaluation is performed by retaining the most significant regions of an image identified by the explanation map and measuring the model's confidence drop (a lower drop is better). 

\subsection{Deletion}
During deletion, we remove the top 20\% values of the SDD CAM by replacing them with 0. We remove the top 20\% values to demonstrate the impact of the most important regions highlighted in the SDD CAM. However, considering the top 20\% values removes more areas than highlighted in the SDD CAM. It's because the extra regions are also included in the top 20\% values of the SDD CAM (though they are not highlighted). 

\begin{figure}[hbt!]
\includegraphics[width=\textwidth]{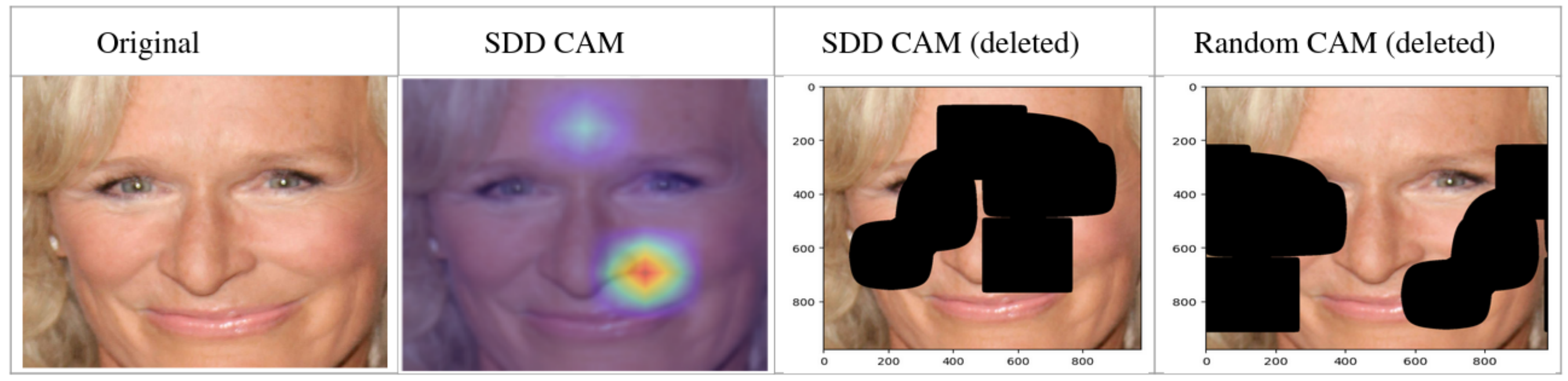}
\caption{An overview of the deletion scheme, which is used for the evaluation of the visual explanation generated by the SDD method. From the left, the first image is the original image, the second image shows the SDD CAM (SDD CAM is overlaid on the image), the top 20\% SDD CAM regions are deleted in the third image, and random CAM regions of the same dimension are deleted in the fourth image.\label{dlt}}
\end{figure}

Fig. \ref{dlt} shows an example of the deletion scheme. To compare with the important regions of the SDD CAM, we remove random areas of the same dimension (we call it random CAM). When certain regions of a test image are removed/covered, the prediction confidence of the face recognition model drops. When SDD CAM regions are removed/covered, the confidence drop should be higher. On the other hand, when random regions are removed/covered, the confidence drop should be lower. 

For example, for a test image, the original confidence of the model is 99\% (for the top class). When SDD CAM regions are removed, the confidence is 70\% (for the original top class). So the confidence drop is 20\%. When random regions are removed, the confidence should be higher than 70\% and the confidence drop should be lower than 20\%. This is because SDD CAM regions are more important for the prediction of the model than random regions. This hypothesis may not apply to all the test images, but it should be true for most of the test images. Therefore, we calculate the average confidence drop for the whole test dataset (4494 images), removing/covering the SDD CAM regions (making the top 20\% values of the SDD CAM 0 as it demonstrates the impact properly). Also, we calculate the average confidence drop after removing/covering random regions of the same dimension. 

\unskip

Another metric to evaluate the SDD CAM results is the change in prediction. In many cases, the prediction of the model changes due to removing/covering certain areas of the test image. For deletion, the prediction change percentage for the whole test dataset should be higher in the case of SDD CAMs. In the case of random CAMs, the prediction change percentage should be lower. For the deletion method, the average confidence drop for SDD CAMs is 62.33\% whereas the average confidence drop for random CAMs is 42.50\%. The prediction change percentage for SDD CAMs is 51.36\% whereas for random CAMs, the prediction change percentage is 29.73\%. The results are shown in Table \ref{tab:dlt}. 
For both evaluation metrics — (1) Average Confidence Drop and (2) Prediction Change Percentage — the explanations generated by SDD CAM outperform those of random CAM. This means that SDD CAM highlights more relevant image regions, leading to a greater drop in model confidence and more frequent prediction changes when those regions are removed/covered.

\begin{table}[H] 
\caption{Comparison of SDD CAM results with those of random CAM using the \emph{deletion method}. The results are generated for the whole test set (4494 images).\label{tab:dlt}}
\centering
\resizebox{\textwidth}{!}{\begin{tabular}{|c|c|c|}
\hline
\textbf{Metric}	& \textbf{SDD CAM}	& \textbf{Random CAM}\\
\hline
Average Confidence Drop (higher is better for deletion)		& 62.33\%		& 42.50\% \\
\hline
Prediction Change Percentage (higher is better for deletion)		& 51.36\% 			& 29.73\% \\
\hline
\end{tabular}}

\end{table}

\subsection{Retention}
In the retention method, only those regions are retained that are removed/covered in the deletion method.

\begin{figure}[hbt!]
\includegraphics[width=15.5 cm]{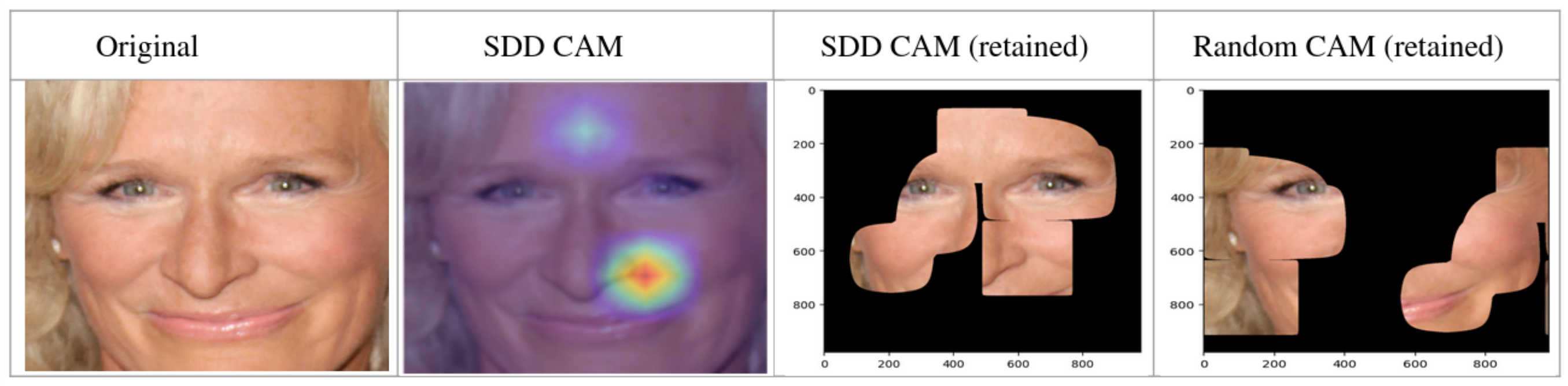}
\caption{An overview of the retention scheme, which is used for the evaluation of the visual explanation generated
by the SDD method. From the left, the first image is the original image, the second image shows the SDD CAM (SDD CAM is overlaid on the image), the top 20\% SDD CAM regions are retained in the third image, and random CAM regions of the same dimension are retained in the fourth image.\label{rtn}}
\end{figure}   
\unskip

As only certain regions are retained, the prediction confidence of the model drops significantly in both the SDD CAM and random CAM cases. However, when the SDD CAM regions are retained, the confidence drop should be lower as the SDD CAM regions are the most relevant regions to the model. When random CAM regions are retained, the confidence drop should be higher, as these regions are not that significant in general. Also, the prediction change percentage should be lower for the SDD CAM than for the random CAM. For SDD CAMs, the average confidence drop for the retention method is 69.43\%. For random CAMs, the average confidence drop is 82.21\% which is higher compared to the average drop for SDD CAMs. The prediction change percentage for SDD CAMs is 57.68\% using the retention method. For random CAMs, the prediction change percentage is 79.93\% which is much higher.  The results are shown in Table \ref{tab:rtn}. 
The evaluation results clearly show that SDD CAM provides more effective explanations compared to random CAM.

\begin{table}[H] 
\caption{Comparison of SDD CAM results with those of random CAM using the \emph{retention method}. The results are generated for the whole test set (4494 images).\label{tab:rtn}}
\centering
\resizebox{\textwidth}{!}{%
\begin{tabular}{|c|c|c|}
\hline
\textbf{Metric}	& \textbf{SDD CAM}	& \textbf{Random CAM}\\
\hline
Average Confidence Drop (lower is better for retention)		& 69.43\%			& 82.21\% \\
\hline
Prediction Change Percentage (lower is better for retention)		& 57.68\% 			& 79.93\% \\
\hline
\end{tabular}}
\end{table}

\section{Discussion}\label{discuss}
The SDD results show that the most relevant regions of a face that a deep learning model considers for making its prediction/decision can be localized very specifically (in a very narrow manner compared to the traditional CAM) using the SDD technique. The SDD explanations are evaluated using the deletion-and-retention scheme (Section \ref{evaluation}). The evaluation results validate that the SDD CAM highlights the most important regions of a test image, which the face recognition model considers for its prediction. For the target class, the face features are determined with respect to other classes. These features are significant to the face recognition model for making its prediction. Such insight can be beneficial for explaining the behavior of deep learning-based face recognition models and increasing their transparency. Also, it will enhance trust in the decisions of the DL-based face recognition systems and can be used to improve their performance.

\section{Conclusion}
This paper visually explains deep learning-based face recognition using an effective localization technique Scaled Directed Divergence. The experiments show that traditional CAMs perform broad localization of relevant face features and are unsuitable in the case of \emph{overlapping classes}. The Scaled Directed Divergence technique can perform very narrow/specific localization of the face features relevant to a deep learning-based face recognition model for making its prediction/decision. The visual explanations generated by the SDD technique are evaluated using the deletion-and-retention scheme. The evaluation results validate the effectiveness of the SDD technique in localizing the most significant face features. These visual explanations will increase the transparency and trustworthiness of existing and new deep learning-based face recognition models. The predictions/decisions of the face recognition models will be more acceptable to the end-users with proper explanations. Also, this can be used to improve the performance of the models by indicating the flaws and biases.

\section{Acknowledgement}
This material is based upon work supported by the Center for Identification Technology Research (CITeR) and the National Science Foundation under Grant No. 1650503.

\bibliographystyle{plain}  
\bibliography{bibtex/bib/ref}   

\begin{thebibliography}{10}

\bibitem{R01}
Housam Khalifa~Bashier Babiker and Randy Goebel.
\newblock An introduction to deep visual explanation.
\newblock {\em arXiv preprint arXiv:1711.09482}, 2017.

\bibitem{R12}
Minchul Kim, Anil~K Jain, and Xiaoming Liu.
\newblock Adaface: Quality adaptive margin for face recognition.
\newblock In {\em Proceedings of the IEEE/CVF conference on computer vision and pattern recognition}, pages 18750--18759, 2022.

\bibitem{R14}
Hong-Wei Ng and Stefan Winkler.
\newblock A data-driven approach to cleaning large face datasets.
\newblock In {\em 2014 IEEE international conference on image processing (ICIP)}, pages 343--347. IEEE, 2014.

\bibitem{R15}
Vitali Petsiuk, Abir Das, and Kate Saenko.
\newblock Rise: Randomized input sampling for explanation of black-box models.
\newblock {\em arXiv preprint arXiv:1806.07421}, 2018.

\bibitem{R8}
Ankit Rajpal, Khushwant Sehra, Rashika Bagri, and Pooja Sikka.
\newblock Xai-fr: explainable ai-based face recognition using deep neural networks.
\newblock {\em Wireless Personal Communications}, 129(1):663--680, 2023.

\bibitem{R3}
Marco~Tulio Ribeiro, Sameer Singh, and Carlos Guestrin.
\newblock " why should i trust you?" explaining the predictions of any classifier.
\newblock In {\em Proceedings of the 22nd ACM SIGKDD international conference on knowledge discovery and data mining}, pages 1135--1144, 2016.

\bibitem{R0}
Wojciech Samek, Thomas Wiegand, and Klaus-Robert M{\"u}ller.
\newblock Explainable artificial intelligence: Understanding, visualizing and interpreting deep learning models.
\newblock {\em arXiv preprint arXiv:1708.08296}, 2017.

\bibitem{R5}
Ramprasaath~R Selvaraju, Michael Cogswell, Abhishek Das, Ramakrishna Vedantam, Devi Parikh, and Dhruv Batra.
\newblock Grad-cam: Visual explanations from deep networks via gradient-based localization.
\newblock In {\em Proceedings of the IEEE international conference on computer vision}, pages 618--626, 2017.

\bibitem{R9}
Avanti Shrikumar, Peyton Greenside, and Anshul Kundaje.
\newblock Learning important features through propagating activation differences.
\newblock In {\em International conference on machine learning}, pages 3145--3153. PMlR, 2017.

\bibitem{R2}
Karen Simonyan, Andrea Vedaldi, and Andrew Zisserman.
\newblock Deep inside convolutional networks: Visualising image classification models and saliency maps.
\newblock {\em arXiv preprint arXiv:1312.6034}, 2013.

\bibitem{R1}
Karen Simonyan and Andrew Zisserman.
\newblock Very deep convolutional networks for large-scale image recognition.
\newblock {\em arXiv preprint arXiv:1409.1556}, 2014.

\bibitem{R10}
Jost~Tobias Springenberg, Alexey Dosovitskiy, Thomas Brox, and Martin Riedmiller.
\newblock Striving for simplicity: The all convolutional net.
\newblock {\em arXiv preprint arXiv:1412.6806}, 2014.

\bibitem{R6}
Edward Verenich, MG~Sarwar Murshed, Nazar Khan, Alvaro Velasquez, and Faraz Hussain.
\newblock Mitigating the class overlap problem in discriminative localization: Covid-19 and pneumonia case study.
\newblock {\em Explainable AI Within the Digital Transformation and Cyber Physical Systems: XAI Methods and Applications}, pages 125--151, 2021.

\bibitem{R7}
Jonathan~R Williford, Brandon~B May, and Jeffrey Byrne.
\newblock Explainable face recognition.
\newblock In {\em European conference on computer vision}, pages 248--263. Springer, 2020.

\bibitem{R4}
Bolei Zhou, Aditya Khosla, Agata Lapedriza, Aude Oliva, and Antonio Torralba.
\newblock Learning deep features for discriminative localization.
\newblock In {\em Proceedings of the IEEE conference on computer vision and pattern recognition}, pages 2921--2929, 2016.

\bibitem{R13}
Zheng Zhu, Guan Huang, Jiankang Deng, Yun Ye, Junjie Huang, Xinze Chen, Jiagang Zhu, Tian Yang, Jiwen Lu, Dalong Du, et~al.
\newblock Webface260m: A benchmark unveiling the power of million-scale deep face recognition.
\newblock In {\em Proceedings of the IEEE/CVF Conference on Computer Vision and Pattern Recognition}, pages 10492--10502, 2021.

\end{thebibliography}

\end{document}